\documentclass[conference]{IEEEtran}
\IEEEoverridecommandlockouts
\usepackage{cite}
\usepackage[dvipsnames]{xcolor}
\usepackage{tikz}
\usepackage{url}
\usepackage{algorithm2e}
\usepackage{hyperref}

\usepackage{subcaption}
\usepackage{booktabs}
\usepackage{comment}
\usepackage{float}
\usepackage{tabularx}
\usepackage{makecell}
\usepackage{pifont}
\newcommand{\cmark}{\ding{51}}%
\newcommand{\xmark}{\ding{55}}%

\usepackage{multirow}
\usepackage{pgfplots}
\usepackage{pgfplotstable}
\usepackage{amsmath,amssymb,amsfonts}

\usepackage{graphicx}
\usepackage{textcomp}
\usepackage{comment}
\usepackage{xcolor}
\def\BibTeX{{\rm B\kern-.05em{\sc i\kern-.025em b}\kern-.08em
    T\kern-.1667em\lower.7ex\hbox{E}\kern-.125emX}}
\begin{document}
\makeatletter
\newcommand{\removelatexerror}{\let\@latex@error\@gobble}
\makeatother

\title{Hybrid-Regularized Magnitude Pruning for Robust Federated Learning under Covariate Shift \\

}
\author{\"Ozg\"u  G\"oksu\\
School of Computing Science\\
University of Glasgow\\
{\tt\small 2718886G@student.gla.ac.uk}
\and
Nicolas Pugeault\\
School of Computing Science\\
University of Glasgow\\
{\tt\small nicolas.pugeault@glasgow.ac.uk}
}

\maketitle

\begin{abstract}
Federated Learning offers a solution for decentralised model training, addressing the difficulties associated with distributed data and privacy in machine learning. However, the fact of data heterogeneity in federated learning frequently hinders the global model's generalisation, leading to low performance and adaptability to unseen data.
This problem is particularly critical for specialised applications such as medical imaging, where both the data and the number of clients are limited. 
In this paper, we empirically demonstrate that inconsistencies in client-side training distributions substantially degrade the performance of federated learning models across multiple benchmark datasets. We propose a novel FL framework using a combination of pruning and regularisation of clients' training to improve the sparsity, redundancy, and robustness of neural connections, and thereby the resilience to model aggregation. 
To address a relatively unexplored dimension of data heterogeneity, we further introduce a novel benchmark dataset, CelebA-Gender, specifically designed to control for within-class distributional shifts across clients based on attribute variations, thereby complementing the predominant focus on inter-class imbalance in prior federated learning research.
Comprehensive experiments on many datasets like CIFAR-10, MNIST, and the newly introduced CelebA-Gender dataset demonstrate that our method consistently outperforms standard FL baselines, yielding more robust and generalizable models in heterogeneous settings.
\end{abstract}


\section{Introduction}
\label{sec:intro}
Federated Learning (FL) offers an efficient paradigm for collaboratively training a shared model across multiple users while preserving client data privacy. Consequently, FL is particularly well-suited for real-world applications where data privacy is critical, such as medical imaging, remote sensing, or processing personal data (eg, faces). FL process is typically initiated by broadcasting an initialised global model from the server to participating clients. Each client independently updates the model using its local data, and the resulting parameters are transmitted back to a central server, where they are aggregated to refine the global model \cite{fedavg}, \cite{fedprox}, \cite{gao2022feddc}. 
In principle, the global model in FL should outperform local models by leveraging diverse client data for improved generalisation. This assumes that local updates can be aggregated effectively to form a stable and performant global model. 
However, non-IID (non-independent and identically distributed) data often induces substantial model drift, leading to divergent local optima and impairing the stability and effectiveness of global aggregation. Addressing this challenge is central to advancing the robustness of federated learning in heterogeneous settings.
We propose \textsc{FedMPR} (\textbf{F}ederated Learning with \textbf{M}agnitude \textbf{P}runing and \textbf{R}egularization), a novel framework designed to enhance federated learning in the presence of data heterogeneity. \textsc{FedMPR} promotes robustness in local models by integrating three key components: (1) magnitude-based pruning to eliminate redundant parameters at the client level; (2) dropout to introduce functional redundancy in decision pathways; and (3) noise injection to regularise model responses and improve resilience to weight perturbations during aggregation. We demonstrate that this framework outperforms standard FL approaches on benchmarks, in particular in datasets with large covariate shifts between clients. 
Additionally, we introduce CelebA-Gender, a novel benchmark dataset for heterogeneous FL. Derived from CelebA~\cite{liu2015deep}, it is specifically designed to evaluate FL methods under challenging conditions where inter-client distribution shifts arise not only from class imbalance, but also from substantial within-class distribution variations.
Client data shift in our study arises from attribute-based gender classification, unlike existing literature \cite{caldas2018leaf}, which typically focuses on examining one attribute at a time for binary classification. CelebA-Gender benchmark provides a more complex data distribution, enabling the evaluation of varying levels of attribute overlap in the image content, ranging from high to low overlap scenarios.

Our framework presents several key contributions:
\begin{itemize}
    \item \textbf{\textsc{FedMPR}:} We propose a novel framework for addressing scenarios with significant covariate shifts across clients' data distributions.  
    \item \textbf{Data Heterogeneity Scenarios:} We evaluate FL approaches under different levels of covariate shift, testing both low and high shift scenarios, including varying numbers of clients (both limited and large populations), as well as imbalanced and limited data, to assess the adaptability and robustness of FL methods across diverse settings.
    \item \textbf{Novel Classification Dataset:} We introduce CelebA-Gender, a reconstructed gender classification dataset derived from attribute-based labels, designed to facilitate a comprehensive evaluation of our framework and enable comparison with existing methods.
\end{itemize}
In FL, cross-silo methods typically involve a smaller number of clients or devices (ranging from 2 to 100), compared to cross-device approaches that can involve millions or even billions of devices \cite{zhu2021federated}. In practice, issues such as data connection loss or slow clients (stragglers) often reduce the effective number of participating clients. This effect is especially pronounced in scenarios with 90$\%$ or more stragglers, resulting in a drastically reduced subset of active clients, sometimes as few as 10, 100, or even fewer. Combined with data heterogeneity, this further impedes model convergence and degrades performance. In this work, we investigate the effects of straggler behaviour, characterised by a limited subset of participating clients, combined with significant data heterogeneity in federated learning. We emphasise the challenges that arise when only a small fraction of clients are active with highly diverse, non-IID data distributions.
\section{Background}
\subsection{Federated Learning}
FL frameworks such as FedAvg \cite{fedavg} typically involve three main steps: \textit{broadcasting}, \textit{local training}, and \textit{model aggregation}. After each communication round, the central server broadcasts the current global model to participating clients. Each client then performs local training on its private data, ensuring data privacy by keeping raw data on-device. Once local updates are completed (e.g., after a fixed number of epochs), clients send their updated model parameters back to the server. The server performs model aggregation, typically by averaging the received parameters, to form a new global model, which is then broadcast in the next round.
Following the typical FL approach, the data $D$ is distributed across $N$ clients and subset of $K$ clients is randomly selected. Let $D_k$ be the local dataset at client $k$, with $n_k = |D_k|$ denoting the number of data points at client $k$. The global objective function in FL is then a weighted average of the local objectives
\begin{equation}
\min _w F(w)=\sum_{k=1}^K \frac{n_k}{n} F_k(w)
\end{equation}
where $n = \sum_{k=1}^{K} n_k$ is the total number of data points across clients and $F_k(w)$ is the local objective function at a client $k$
\begin{equation}
 \quad F_k(w)=\frac{1}{n_k} \sum_{i \in \mathcal{D}_k} f_i(w)
\end{equation}
Current research predominantly relies on FedAvg-based model aggregation and training procedures \cite{oh2021fedbabu}, \cite{tamirisa2024fedselect} \cite{li2021model} that can broadly be categorised into two main directions: enhancing local training and refining the model aggregation strategy. Our work aligns with the first direction, targeting the local training phase. By improving local optimisation, we aim to reduce divergence across client models, thereby indirectly mitigating challenges typically addressed at the aggregation stage. 
\begin{figure*}
    \centering
    \includegraphics[width=0.85\linewidth]{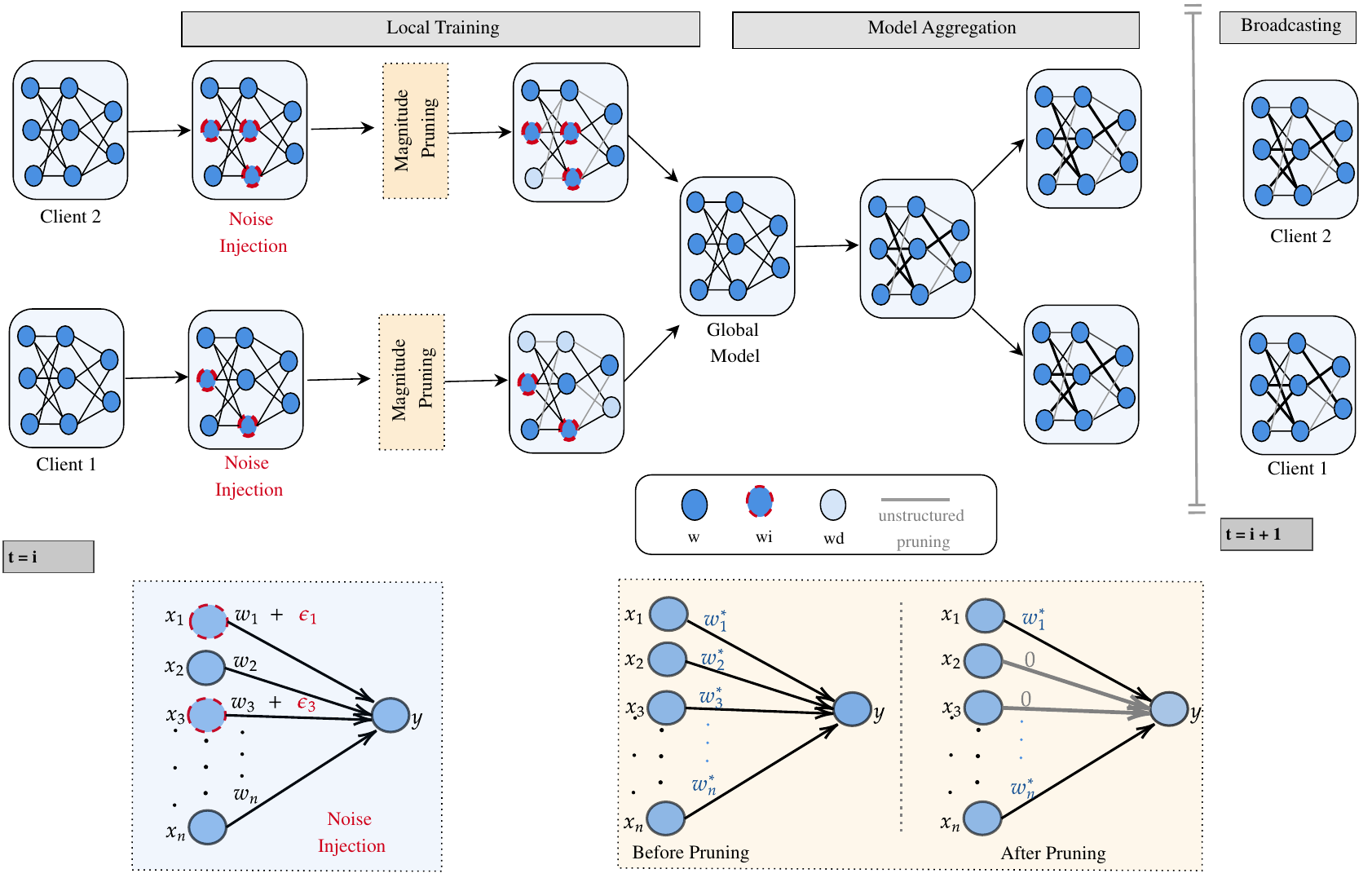}
    \caption{\textsc{FedMPR} framework, each client applies a tailored regularisation that combines dropout during forward passes with Gaussian noise injection inside each basic block. At the $i$th iteration, the central server broadcasts the global model weights to clients, which then perform local training before model aggregation. Specifically, $w$ denotes the original weights, $wi$ the weights perturbed by Gaussian noise, and $wd$ the zeroed (pruned) weights.}
    \label{fig:method}
\end{figure*}
\subsection{Related Work}
\textbf{Federated Parameter Selection }
FL approaches encounter several critical challenges, particularly poor convergence on highly heterogeneous data and the lack of solutions for individual clients. To address these issues, parameter selection or decoupling FL approaches introduce tailored models for heterogeneity \cite{oh2021fedbabu}, \cite{tamirisa2024fedselect}, \cite{tan2022towards}, \cite{jia2024dapperfl}. Among these approaches, some studies \cite{oh2021fedbabu}, \cite{tamirisa2024fedselect} manually partition the model into personalised and shared parameters. 
Personalised methods can address the data heterogeneity by tailoring models to individual clients. However, when clients have imbalanced or highly distributed data, these algorithms may struggle to learn robust features. Moreover, personalized approaches typically do not yield a unified global model. Therefore a significant question remains unresolved: how to effectively eliminate redundant parameters in local models, especially when there are relatively the contribution of few clients and substantial data heterogeneity among them, while training a global model without personalization.\\

\textbf{Covariate Shift in FL}
Numerous methods have been proposed to address the challenges posed by data heterogeneity in FL  \cite{luo2021no}, \cite{dai2023tackling}, \cite{huang2023rethinking}. These approaches often introduce proximal terms to constrain local updates concerning the global model, aiming to mitigate the adverse effects of client data diversity. 
While these approaches reduce the divergence between local and global models, they simultaneously hinder local convergence, thereby diminishing the amount of information gained in each communication round. 
Unfortunately, many existing FL algorithms fail to consistently cause stable performance improvements across diverse non-IID scenarios, when compared to traditional baselines \cite{li2022federated}, \cite{fedavg}, \cite{gao2022feddc}, especially in vision tasks such as the classification of genders or emotions. For instance, FedBABU \cite{oh2021fedbabu} addresses data heterogeneity by updating only the model's body parameters while keeping the head randomly initialised and fixed during local training, sharing only the body with the central server. 
However, a significant gap remains in effectively addressing data heterogeneity, particularly in cases where covariate shifts among clients are more substantial. To improve FL robustness to non-IID cases, PFL approaches have shown that personalising specific layers can be an effective approach~\cite{collins2021exploiting}. However, over-personalised client models may overfit and undermine the generalisation benefits of a global model, when the distribution variances across clients become excessively significant (high label distribution skew).  \\

\textbf{Regularisation in FL}
FedProx\cite{fedprox} introduces a proximal term in the loss function that regularises the local model updates. This regularisation term limits the local updates from diverging too far from the global model, addressing issues such as the partial participation of clients in FL. This work focuses on scenarios with up to 90$\%$ stragglers across various datasets, resulting in a minimum of approximately 20 active clients. However, even in federated learning settings with a large number of clients, many of which experience slow training or communication delays. Existing works do not adequately address or explain these challenges.
SCAFFOLD \cite{karimireddy2019scaffold} introduces a control variate-based regularisation method to correct for client covariate shifts in non-IID. The work \cite{lee2024regularizing} estimates weight distribution regularisation for each client using the FedAvg under 100 clients maximum.
Consequently, the regularisation-based FL approaches \cite{wen2022federated} may suffer from real-world applications where the classes are completely different per client, leading to instability in distribution estimation. MOON \cite{li2021model} reformulates contrastive loss to leverage the global model, which captures representations from the entire data distribution. It builds on a similar principle as FedProx by constraining local updates concerning the global model.
In contrast, our proposed method trains each client model and eliminates redundant parameters, thereby reducing the impact of updated parameters on local training and global model aggregation, leading to better convergence.
\section{Methodology}
\subsection{FedMPR}
\label{Methodology}
Unlike Personalised Federated Learning (PFL), which allows clients to customise or fine-tune a local model based on its data, our approach follows the standard FL framework. In our approach, clients train the same shared global model without making any modifications or changes. 
Each client only applies the global model as is, without any modifications to enhance it to fit its local data. We reformulate Iterative Magnitude Pruning (IMP) \cite{frankle2018lottery} (Algorithm~\ref{alg2}) by introducing a round-wise strategy, enabling progressive sparsification across rounds instead of applying it only once per round.
Our approach enhances the standard FL framework by mitigating the adverse effects of model aggregation. As illustrated in Figure \ref{fig:method} (two-client example), and Algorithm \ref{alg1} the proposed method introduces three key components: \\
\textbf{Pruning:} In neural networks, weights with small magnitudes often contribute minimally to the model’s output \cite{han2015learning}. While such parameters may have a negligible impact in centralised training, in FL, they can amplify aggregation misalignment under data heterogeneity. \\
\textbf{Redundancy:} Dropout applied during local training induces redundancy by encouraging diverse subnetworks \cite{wen2022federated}, which can improve the robustness of model aggregation under non-IID conditions by mitigating client-specific overfitting and reducing the variance in local updates. \\
\textbf{Robustness:} Injecting noise during local training enhances robustness to small weight perturbations introduced during aggregation. Additionally, applying dropout promotes redundancy in neural pathways, further improving the model’s tolerance to aggregation noise. Together, these techniques help mitigate client-side overfitting in federated settings. 
\begin{figure}[h]
\centering
\scriptsize
\begin{minipage}[t]{0.48\textwidth}
\removelatexerror
\begin{algorithm}[H]
\caption{Federated Magnitude Pruning and Regularisation (\textsc{FedMPR})}
\label{alg1}
\scriptsize
\KwIn{Number of clients $n$; total rounds $T$; pruning frequency $f$; threshold $\theta$; initial global model $w_0$; prune percentage $p$; sparsity $\beta$}
\KwOut{Final global model $w_T$}
Partition $\mathcal{D}$ into $\{\mathcal{D}_1, \mathcal{D}_2, \ldots, \mathcal{D}_n\}$ \\
\For{each client $c_i \in \{1, \ldots, n\}$}{
  Initialize $c_i$ with $w_0$ \\
  Initialize $m_i$ masks
}
\For{each round $t = 1$ \KwTo $T$}{
  Server broadcasts $w_{t-1}$ to clients $c \in \mathcal{C}_t$ \\
  \ForEach{client $c \in \mathcal{C}_t$ in parallel}{
    Apply mask: $w_c^t \leftarrow m_c \odot w_{t-1}$ \\
    Train $w_c^t$ on $\mathcal{D}_c$ \\
    \If{sparsity of $w_c^t > \beta$}{
      $w_c^t \leftarrow \text{PruneWeights}(w_c^t, p)$
    }
  }
  Clients send $w_c^t$ to server \\
  Server aggregates: $w_t \leftarrow \sum_{c \in \mathcal{C}_t} \frac{n_c}{n_{\mathcal{C}_t}} w_c^t$
}
\Return{$w_T$}
\vspace{2mm}
\end{algorithm}

\end{minipage}
\hfill
\vspace{2mm}
\begin{minipage}[t]{0.48\textwidth}
\removelatexerror
\begin{algorithm}[H]
\caption{PruneWeights: Iterative Magnitude Pruning}
\label{alg2}
\scriptsize
\KwIn{Model parameters $W$, prune percentage $p$}
\KwOut{Pruned model parameters $W'$}
\ForEach{parameter tensor $w$ in $W$}{
  \If{$w$ is conv or linear weights}{
    $T \leftarrow$ flatten($w$) \\
    $A \leftarrow |T|$ \\
    $thr \leftarrow$ percentile($A$, $p \times 100$) \tcp*{Threshold}

    \ForEach{$a_i$ in $A$}{
      \eIf{$a_i > thr$}{$mask_i \leftarrow 1$}{$mask_i \leftarrow 0$}
    }
    $w' \leftarrow T \odot mask$ \tcp*{Element-wise} 
    reshape $w'$ to shape of $w$ \\ replace $w$ in $W$ with $w'$
  }
}
\Return{$W'$}
\vspace{2mm}
\end{algorithm}
\end{minipage}
\end{figure}
\begin{figure*}[t]
    \centering
    \includegraphics[width=1.\textwidth]{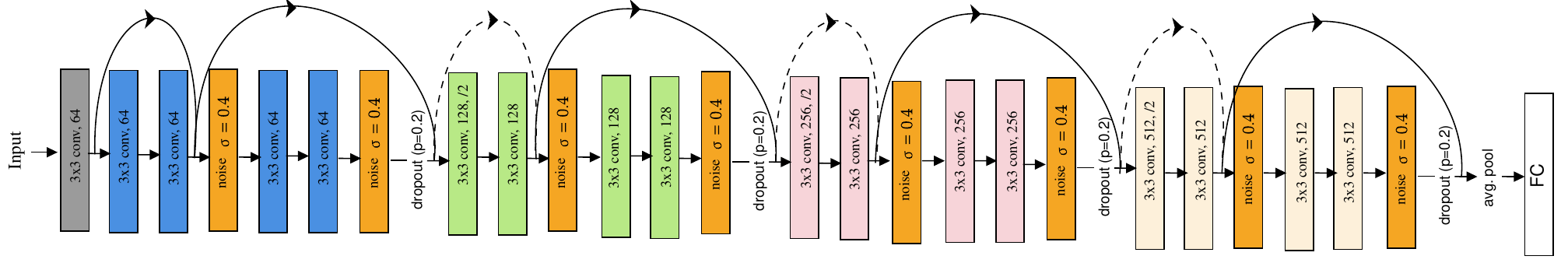}
    \caption{Configured ResNet-18 architecture with regularisation layers integration.}
    \label{fig:r18model}
\end{figure*}
\subsection{ FL with Varied Data Distributions}
Covariate shift, a central form of data heterogeneity in FL, arises from variations in input feature distributions across clients. While the Dirichlet distribution is commonly used to simulate non-IID conditions \cite{tamirisa2024fedselect, oh2021fedbabu}, it introduces both class imbalance and data overlap. In this work, we explore alternative partitioning strategies to model covariate shifts more explicitly and also consider scenarios with highly similar client distributions.\\
\textbf{Dirichlet Distribution:} This distribution allows for controlled variation in data partitioning, provides many heterogeneous client data scenarios by $\alpha$ parameter, which controls the imbalance across clients like previous studies \cite{tamirisa2024fedselect, li2021model}. Smaller $\alpha$ values like 0.1, lead to more skewed, imbalanced, non-IID data partitions. In our experiments, we use 0.1 and 0.5 with 10 and 100 clients.\\
\textbf{Low versus High Covariate Shift conditions:}
One limitation of using the Dirichlet distribution to control the client data distribution is that the training sets are not mutually exclusive, and therefore multiple clients will see the same training examples. This is unlike any real scenario where each client would hold completely distinct data. Therefore, in addition to the Dirichlet distribution, we experimented with mutually exclusive client data in two conditions, denoted as low covariate shift (low-CS) and high covariate shift (high-CS) in the paper. 
In the low-CS condition, all clients receive an equal number of training examples from each target class, where each example is only allocated to one client. 
In the high-CS condition on the other hand, each client receives a subset of classes, for example in a two-client setup on MNIST, client A would train on examples of the classes $y_1 \in Y_1=\left\{ 0,1,2,3,4\right\}$, and client B on $y_2 \in Y_2=\left\{5,6,7,8,9\right\}$. 

\subsection{CelebA-Gender Dataset} We introduce CelebA-Gender, a novel gender classification dataset derived from CelebA, designed to evaluate data heterogeneity arising from within-class distribution differences rather than relying on a single attribute for class definition. To the best of our knowledge, existing FL literature defines classes based on one attribute, whereas each image inherently contains multiple attributes. The overlap and non-overlap of these attributes create more specific and complex data distribution patterns, enabling a more realistic assessment of heterogeneity. The dataset consists of approximately 40K images at a resolution of $178 \times 218$ under female/male classes. Dataset reconstructed with five attributes, which are \textit{Black Hair, Smiling, High Cheekbones, Attractive, Mouth Slightly Open}. While multiple attributes can be defined, we focus on five to ensure sufficient overlap per sample. Increasing the number of target attributes reduces the likelihood of their co-occurrence, shrinking the sample space and limiting the effectiveness of federated learning due to data scarcity. \\
We assessed the similarity of the images in each dataset using the FID (Fréchet Inception Distance) \cite{heusel2017gans}, and CMMD (CLIP embeddings with Maximum Mean Discrepancy) \cite{jayasumana2024rethinking}. Lower values for FID and CMMD scores represent higher similarity between data distributions. 
\begin{figure}[H]
    \centering
    \begin{minipage}[t]{0.48\textwidth}
        \centering
        \scriptsize
        \textbf{(a) Mutually Exclusive Attributes} \\
        \includegraphics[width=0.18\linewidth]{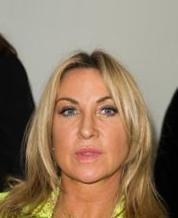}
        \includegraphics[width=0.18\linewidth]{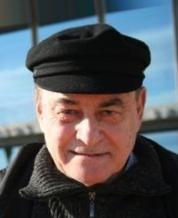}
        \includegraphics[width=0.18\linewidth]{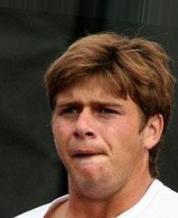}
        \includegraphics[width=0.18\linewidth]{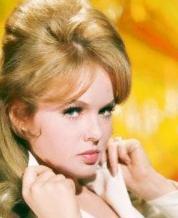}
        \includegraphics[width=0.18\linewidth]{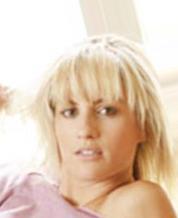} \\
        \vspace{1mm}
        \footnotesize{Each sample possesses only one specified attribute.}
        \label{fig:dataceleb5a}
    \end{minipage}
    \hfill
    \vspace{3mm}
    \begin{minipage}[t]{0.48\textwidth}
        \centering
        \scriptsize
        \textbf{(b) Mutually Inclusive Attributes} \\
        \includegraphics[width=0.18\linewidth]{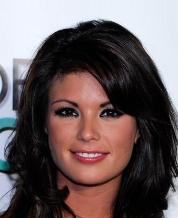}
        \includegraphics[width=0.18\linewidth]{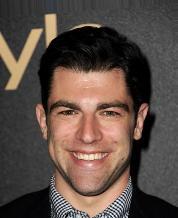}
        \includegraphics[width=0.18\linewidth]{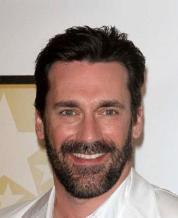}
        \includegraphics[width=0.18\linewidth]{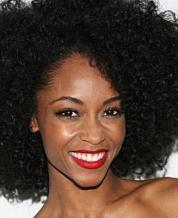}
        \includegraphics[width=0.18\linewidth]{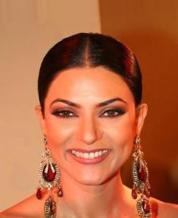} \\
        \vspace{1mm}
        \footnotesize{Each sample possesses all five specified attributes.}
        \label{fig:dataceleb5b}
    \end{minipage}
    \caption{Samples from the CelebA-Gender dataset (5 attributes) illustrating mutually exclusive (a) and inclusive (b) attribute combinations across gender classes.} 
    \label{fig:dataceleb_combined}
\end{figure} 
\textbf{Mutually Exclusive} Each sample has only one of the target attributes, and never all together, leading to high-CS. In Figure~\ref{fig:dataceleb_combined} (a), samples show high cheekbones but lack the other four attributes. \\
\textbf{Mutually Inclusive} Each sample contains all target attributes simultaneously, resulting in a low-CS. In Figure~\ref{fig:dataceleb_combined} (b), every sample exhibits five attributes.\\

\begin{figure}[H]
    \centering
    \begin{minipage}[t]{0.48\textwidth}
        \centering
        \scriptsize
        \textbf{(a) Mutually Exclusive Attributes} \\
        \includegraphics[width=0.18\linewidth]{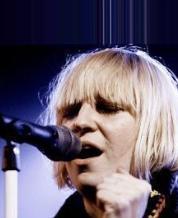}
        \includegraphics[width=0.18\linewidth]{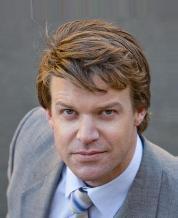}
        \includegraphics[width=0.18\linewidth]{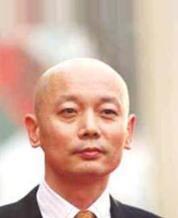}
        \includegraphics[width=0.18\linewidth]{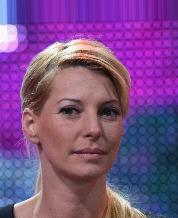}
        \includegraphics[width=0.18\linewidth]{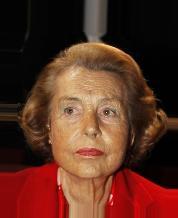} \\
        \vspace{1mm}
        \footnotesize{Each sample possesses only one specified attribute.}
    \end{minipage}
    \hfill
    \vspace{3mm}
    \begin{minipage}[t]{0.48\textwidth}
        \centering
        \scriptsize
        \textbf{(b) Mutually Inclusive Attributes} \\
        \includegraphics[width=0.18\linewidth]{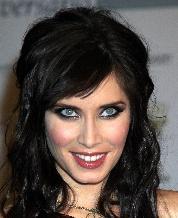}
        \includegraphics[width=0.18\linewidth]{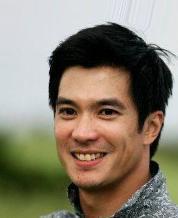}
        \includegraphics[width=0.18\linewidth]{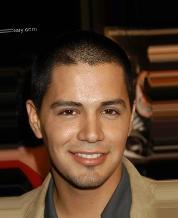}
        \includegraphics[width=0.18\linewidth]{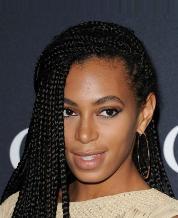}
        \includegraphics[width=0.18\linewidth]{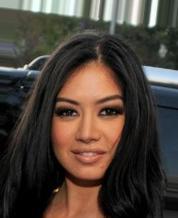} \\
        \vspace{1mm}
        \footnotesize{Each sample possesses all seven specified attributes.}
    \end{minipage}
    \caption{Samples from the CelebA-Gender dataset (7 attributes) illustrating mutually exclusive (a) and inclusive (b) attribute combinations across gender classes.} 
    \label{fig:7att}
\end{figure}

As the number of attributes increases to seven for the CelebA-Gender dataset, as shown in Figure \ref{fig:7att}, the similarity among samples also increases. Increasing the number of attributes in the CelebA-Gender dataset effectively reduces the sample selection space, causing mutually inclusive or exclusive cases to become more similar to each other. This phenomenon is evident as we vary the attribute count from 3 up to 7.
\subsection{Experimental Setup} 
\textbf{Model $\And$ Datasets} \\ 
We use ResNet-18 as the backbone architecture, initialising both global and local models randomly. As shown in Figure \ref{fig:r18model}, we modify the standard ResNet-18 by incorporating noise injection (percentage 0.4), layers and dropout (ratio 0.2) within each basic block. For small-scale datasets such as CIFAR-10 \cite{krizhevsky2009learning}, MNIST \cite{lecun1998gradient}, Fashion-MNIST \cite{xiao2017fashion}, and SVHN \cite{netzer2011reading}, we reduce the convolutional kernel size within the basic blocks to $3\times3$ to better fit the lower-resolution inputs. For larger-scale datasets such as CelebA-Gender \footnote{\url{https://anonymous.4open.science/r/Dataset-E468/README.md}}. and RAF-DB \cite{li2017reliable}, we retain the original kernel sizes.\\
\textbf{Training Setup} \\
We set the number of clients to 2 for both low and high covariate shift scenarios, allowing us to evaluate the performance with limited clients exhibiting higher similarity in their data, as well as vice versa. Additionally, we use the Dirichlet distribution with $\alpha=0.1, 0.5$ for a higher number of clients (10 and 100). We employed Stochastic Gradient Descent (SGD) and SGD momentum is 0.9, a batch size is 128 to train each model, using a learning rate of 2e-2. Each local model was trained for 5 epochs per round, communication round is 100.
\section{Results}
Pruning methods with Lottery Ticket Hypothesis (LTH) perform better on many real-world applications, like PruneFL \cite{jiang2022model}, LotteryFL \cite{li2020lotteryfl} and FedSelect\cite{tamirisa2024fedselect}. Among them, we focus on FedSelect for comparison, as it incorporates a gradient-driven LTH strategy tailored for personalisation closely aligning with the sparsity mechanisms employed in our method. This enables a principled evaluation of selective parameter training within the FL framework. \\
In Table \ref{tab:low_high_CS}, we present the performance  FL algorithms on a range of benchmarks, in the low-CS (as evidenced by low FID, CMMD scores between the two local data distributions). When local datasets exhibit high similarity, all methods achieve around $50\%$ top-1 accuracy across several benchmarks. However, as inter-client dissimilarity increases, performance degrades significantly due to the absence of overlap between local data, which challenges standard model aggregation methods. In contrast, pruning-based approaches such as FedSelect and \textsc{FedMPR} demonstrate improved robustness by extracting more generalizable features from client data. \\
As shown in Table \ref{tab:low_high_CS}(b), they achieve notably higher accuracy, particularly on the CIFAR10 and CelebA-Gender datasets. \textsc{FedMPR} consistently outperforms baselines across datasets, indicating that leveraging model redundancy enhances performance under both balanced and imbalanced data distributions.\\ 
Across data scenarios with increasing covariate shift, model aggregation methods tend to degrade under high imbalance. In contrast, approaches incorporating parameter pruning and overfitting mitigation enable clients to perform strongly across diverse datasets.\\
\begin{table*}[htbp]
\centering
\scriptsize
\begin{minipage}[t]{\textwidth}
\centering
\begin{tabular}{lllllll}
\textbf{Method}  & \textbf{CIFAR10}    & \textbf{MNIST}  & \textbf{FMNIST}  & \textbf{SVHN}  & \textbf{CelebA-Gender}  & \textbf{RAF-DB}   \\ [0.05cm] \hline \hline
Supervised & \textbf{91.51} & \textbf{99.08} & \textbf{94.18 }& \textbf{93.89} & \textbf{98.39} & \textbf{77.80}   \\ 
FedAvg & 77.00{\color{gray}$\pm $0.33 } & 99.01{\color{gray}$\pm $0.04}  &  90.73{\color{gray}$\pm$0.31} & 91.11{\color{gray}$\pm $0.37 } & 91.18{\color{gray}$\pm $8.23 } & 43.42{\color{gray}$ \pm$1.78}  \\ 
FedProx & 76.45{\color{gray}$\pm $0.89 } & 98.94{\color{gray}$\pm $0.16 } & 91.31{\color{gray}$\pm $0.54 } & 91.07{\color{gray}$\pm $0.17 } & 72.05{\color{gray}$ \pm$3.86}  & 64.31{\color{gray}$ \pm$1.45}  \\ 
FedDC & 74.37{\color{gray}$\pm $2.97 }  & 98.58{\color{gray}$\pm $0.11 } & 90.04{\color{gray}$\pm $0.61 } & 90.44{\color{gray}$\pm $0.43 } & 61.67{\color{gray}$ \pm$7.97} & 62.43{\color{gray}$ \pm$1.18} \\ 
FedDyn & 77.72{\color{gray}$\pm $0.65 }  & 98.98{\color{gray}$\pm $0.20 } & 91.13{\color{gray}$ \pm$1.16 } & 91.57{\color{gray}$\pm $0.10 } & 72.07{\color{gray}$ \pm$4.13} & 64.39{\color{gray}$ \pm$1.91}  \\ 
SCAFFOLD & 76.73{\color{gray}$\pm $0.28 } & 98.77{\color{gray}$\pm $0.25 } & 90.87{\color{gray}$\pm $0.39 } & 90.97{\color{gray}$\pm $0.05 } & 72.13{\color{gray}$ \pm$4.12} & 64.51{\color{gray}$ \pm$1.59} \\ 
FedSelect & 76.25{\color{gray}$\pm $1.17 }  & 94.91{\color{gray}$\pm $2.16 } & 89.78{\color{gray}$\pm $2.58 } & 90.32{\color{gray}$\pm $1.16 } & 93.14{\color{gray}$ \pm$3.64} & 72.74{\color{gray}$ \pm$1.48}  \\ 
\textsc{FedMPR} & \textbf{86.61{\color{gray}$\pm $1.26 }} & \textbf{99.49{\color{gray}$\pm $0.07 }} & \textbf{93.10{\color{gray}$\pm $0.46 }} & \textbf{94.17{\color{gray}$\pm $0.21 }} & \textbf{96.93{\color{gray}$ \pm$ 1.05}} & \textbf{74.92{\color{gray}$ \pm$1.16}}  \\ [0.1cm] \hline
FID Score & 1.68 &  0.63 & 0.83 & 0.52 & 12.00 & 7.79  \\ \hline
CMMD Score & $\sim$0.00 & $\sim$0.00 & $\sim$0.00 & $\sim$0.00 & $\sim$0.00 & $\sim$0.00 \\ \hline 
\end{tabular}
\vspace{2mm} \\
{\textbf{(a)} FL accuracy on \textit{low-CS} condition for 2 clients. Lower FID or CMMD suggests higher data similarity between clients.}
\end{minipage}

\vspace{3mm} 

\begin{minipage}[t]{\textwidth}
\centering
\begin{tabular}{lllllll}
\textbf{Method}    & \textbf{CIFAR10}  &  \textbf{MNIST}  & \textbf{FMNIST}  & \textbf{SVHN} & \textbf{CelebA-Gender}  & \textbf{RAF-DB}  \\ [0.05cm] \hline \hline
Supervised & \textbf{91.51} &\textbf{99.08} & \textbf{94.18 }& \textbf{93.89} & \textbf{98.39} & \textbf{77.80}  \\  
FedAvg & 51.28{\color{gray}$\pm $1.55 } & 89.84{\color{gray}$\pm $1.31}  &  80.69{\color{gray}$\pm $1.30 } & 75.69{\color{gray}$\pm $3.41 } & 47.82{\color{gray}$ \pm$1.93} & 45.06{\color{gray}$ \pm$1.39} \\ 
FedProx & 54.71{\color{gray}$\pm $1.25 } & 87.28{\color{gray}$\pm $3.04 } & 81.49{\color{gray}$\pm $1.18 } & 74.90{\color{gray}$\pm $1.21 } & 71.80{\color{gray}$ \pm$1.92} & 45.99{\color{gray}$ \pm$2.84} \\ 
FedDC & 41.09{\color{gray}$\pm $1.55 }  & 65.62{\color{gray}$\pm $9.79 } & 60.11{\color{gray}$\pm $6.75 } & 59.02{\color{gray}$\pm $2.47 } & 62.77{\color{gray}$ \pm$6.10} & 40.55{\color{gray}$ \pm$0.82} \\ 
FedDyn & 49.48{\color{gray}$\pm $0.72 } & 85.53{\color{gray}$\pm $2.05 } & 79.33{\color{gray}$\pm $2.55 } & 69.71{\color{gray}$\pm $2.77 } & 70.50{\color{gray}$ \pm$3.16} & 43.12{\color{gray}$ \pm$1.26} \\ 
SCAFFOLD & 49.18{\color{gray}$\pm $0.93 }  & 78.73{\color{gray}$\pm $5.59 } & 71.06{\color{gray}$\pm $3.12 } & 59.09{\color{gray}$\pm $2.26 } & 67.63{\color{gray}$ \pm$7.16} & 43.11{\color{gray}$ \pm$1.27}\\ 
FedSelect & 61.97{\color{gray}$\pm $5.96 }  & 91.02{\color{gray}$\pm $4.76 } & 85.01{\color{gray}$\pm $5.16 } & 81.00{\color{gray}$\pm$4.39 } & 52.50{\color{gray}$ \pm$ 4.40} & 67.45{\color{gray}$ \pm$4.38} \\ 
\textsc{FedMPR} & \textbf{75.22{\color{gray}$\pm $5.35 }} & \textbf{98.99{\color{gray}$\pm $0.34}} & \textbf{88.92{\color{gray}$\pm $2.24 }} & \textbf{88.24{\color{gray}$\pm $4.67 }} & \textbf{84.23{\color{gray}$ \pm$1.22}} & \textbf{69.41{\color{gray}$ \pm$5.54}}  \\[0.1cm] \hline 
FID Score & 70.69 &  43.14 & 47.49 &4.52 & 82.85 & 9.88  \\ \hline 
CMMD Score & 0.16 &  0.20 & 0.10 & 0.12 & 1.52 &0.17  \\ \hline 
\end{tabular}
\vspace{2mm} \\
{\textbf{(b)} FL accuracy on \textit{high-CS} condition for 2 clients. Higher FID and CMMD values indicate greater data heterogeneity.}
\end{minipage}
\caption{Comparison of global model accuracy (\%) under (a) low and (b) high-CS settings. Average test accuracy over three runs.  ($\pm$) denotes the standard deviation from the average.}
\label{tab:low_high_CS}
\end{table*}
\begin{figure*}[ht]
    \centering
    \tiny
    \begin{minipage}[t]{0.45\textwidth}
        \centering
        \begin{tikzpicture}
        \begin{axis}[
            xlabel={\tiny{Number of Samples}},
            ylabel={\tiny{Accuracy}},
            grid=major,
            width=\textwidth, 
            height=4.5cm,
            ymin=0, ymax=1, 
            xmode=normal, 
            xtick={100, 500, 1000, 2000, 3000, 5000},
            xticklabels={\tiny 100, \tiny 500, \tiny1000, \tiny2000, \tiny3000, \tiny5000},
            xticklabel style={rotate=42, anchor=east}, 
            enlargelimits=0.1, 
            scaled ticks=false, 
             legend pos=outer north east, 
        ]
            \addplot+[
                mark=square*, color=blue,
                error bars/.cd, y dir=both, y explicit
            ]
            coordinates {
                (100, 0.34) +- (0,0.05)
                (500, 0.58) +- (0,0.04)
                (1000, 0.65) +- (0,0.04)
                (2000, 0.76) +- (0,0.02)
                (3000, 0.81) +- (0,0.01)
                (5000, 0.87) +- (0,0.01)
            };

            \addplot+[
                mark=o, color=red,
                error bars/.cd, y dir=both, y explicit
            ]
            coordinates {
                (100, 0.10) +- (0,0.01)
                (500, 0.38) +- (0,0.05)
                (1000, 0.45) +- (0,0.05)
                (2000, 0.58) +- (0,0.05)
                (3000, 0.68) +- (0,0.02)
                (5000, 0.76) +- (0,0.01)
            };

            \addplot+[
                mark=triangle*, color=green,
                error bars/.cd, y dir=both, y explicit
            ]
            coordinates {
                (100, 0.39) +- (0,0.02)
                (500, 0.48) +- (0,0.03)
                (1000, 0.61) +- (0,0.03)
                (2000, 0.67) +- (0,0.01)
                (3000, 0.71) +- (0,0.03)
                (5000, 0.74) +- (0,0.03)
            };

            \addplot+[
                mark=star, color=orange,
                error bars/.cd, y dir=both, y explicit
            ]
            coordinates {
                (100, 0.44) +- (0,0.02)
                (500, 0.55) +- (0,0.02)
                (1000, 0.64) +- (0,0.005)
                (2000, 0.71) +- (0,0.01)
                (3000, 0.74) +- (0,0.01)
                (5000, 0.78) +- (0,0.01)
            };

            \addplot+[
                mark=x, color=purple,
                error bars/.cd, y dir=both, y explicit
            ]
            coordinates {
                (100, 0.37) +- (0,0.02)   
                (500, 0.50) +- (0,0.02)
                (1000, 0.58) +- (0,0.02)
                (2000, 0.68) +- (0,0.01)
                (3000, 0.72) +- (0,0.01)
                (5000, 0.77) +- (0,0.02)
            };

            \addplot+[
                mark=diamond*, color=cyan,
                error bars/.cd, y dir=both, y explicit
            ]
            coordinates {
                (100, 0.38) +- (0,0.03) 
                (500, 0.51) +- (0,0.03)
                (1000, 0.60) +- (0,0.01)
                (2000, 0.67) +- (0,0.01)
                (3000, 0.73) +- (0,0.01)
                (5000, 0.77) +- (0,0.01)
            };

            \addplot+[
                mark=triangle, color=brown,
                error bars/.cd, y dir=both, y explicit
            ]
            coordinates {
                (100, 0.38) +- (0,0.02)
                (500, 0.52) +- (0,0.01)
                (1000, 0.60) +- (0,0.01)
                (2000, 0.67) +- (0,0.03)
                (3000, 0.69) +- (0,0.02)
                (5000, 0.76) +- (0,0.01)
            };
        \end{axis}
        \end{tikzpicture} \\
        {\hspace{0.5cm} \scriptsize Low-CS}
       
    \end{minipage}
    \hfill
    \begin{minipage}[t]{0.45\textwidth}
        \centering
        \begin{tikzpicture}
        \begin{axis}[
            xlabel={\tiny{Number of Samples}},
            ylabel={\tiny{Accuracy}},
            grid=major,
            width=\textwidth, 
            height=4.5cm,
            ymin=0, ymax=1, 
            xmode=normal, 
            xtick={100, 500, 1000, 2000, 3000, 5000},
           xticklabels={\tiny 100, \tiny 500, \tiny1000, \tiny2000, \tiny3000, \tiny5000},
            xticklabel style={rotate=42, anchor=east}, 
            enlargelimits=0.1, 
            scaled ticks=false, 
             legend pos=outer north east, 
        ]
            \addplot+[
                mark=square*, color=blue,
                error bars/.cd, y dir=both, y explicit
            ]
            coordinates {
                (100, 0.17) +- (0,0.02)
                (500, 0.38) +- (0,0.04)
                (1000, 0.44) +- (0,0.09)
                (2000, 0.54) +- (0,0.05)
                (3000, 0.60) +- (0,0.05)
                (5000, 0.75) +- (0,0.05)
            };

            \addplot+[
                mark=o, color=red,
                error bars/.cd, y dir=both, y explicit
            ]
            coordinates {
                (100, 0.10) +- (0,0.002)
                (500, 0.30) +- (0,0.03)
                (1000, 0.35) +- (0,0.08)
                (2000, 0.47) +- (0,0.03)
                (3000, 0.54) +- (0,0.06)
                (5000, 0.62) +- (0,0.06)
            };

            \addplot+[
                mark=triangle*, color=green,
                error bars/.cd, y dir=both, y explicit
            ]
            coordinates {
                (100, 0.23) +- (0,0.04)
                (500, 0.34) +- (0,0.04)
                (1000, 0.35) +- (0,0.02)
                (2000, 0.40) +- (0,0.01)
                (3000, 0.42) +- (0,0.01)
                (5000, 0.42) +- (0,0.02)
            };

            \addplot+[
                mark=star, color=orange,
                error bars/.cd, y dir=both, y explicit
            ]
            coordinates {
                (100, 0.27) +- (0,0.0)
                (500, 0.33) +- (0,0.01)
                (1000, 0.34) +- (0,0.01)
                (2000, 0.36) +- (0,0.01)
               
                (3000, 0.41) +- (0,0.01)
                (5000, 0.50) +- (0,0.01)
    };
        \addplot+[
            mark=x, color=purple,
            error bars/.cd, y dir=both, y explicit
        ]
        coordinates {
            (100, 0.18) +- (0,0.02)   
            (500, 0.35) +- (0,0.02)
            (1000, 0.39) +- (0,0.04)
            (2000, 0.45) +- (0,0.02)
            (3000, 0.51) +- (0,0.02)
            (5000, 0.56) +- (0,0.02)
        };

        \addplot+[
            mark=diamond*, color=cyan,
            error bars/.cd, y dir=both, y explicit
        ]
        coordinates {
            (100, 0.22) +- (0,0.04) 
            (500, 0.30) +- (0,0.03)
            (1000, 0.35) +- (0,0.05)
            (2000, 0.40) +- (0,0.02)
            (3000, 0.45) +- (0,0.01)
            (5000, 0.50) +- (0,0.02)
        };

        \addplot+[
            mark=triangle, color=brown,
            error bars/.cd, y dir=both, y explicit
        ]
        coordinates {
            (100, 0.25) +- (0,0.02)
            (500, 0.33) +- (0,0.03)
            (1000, 0.39) +- (0,0.05)
            (2000, 0.46) +- (0,0.04)
            (3000, 0.53) +- (0,0.02)
            (5000, 0.57) +- (0,0.02)
        };
    \end{axis}
    \end{tikzpicture} \\
    {\hspace{0.5cm} \scriptsize High-CS}
    
\end{minipage}

    \begin{tikzpicture}
\begin{axis}[
    width=\textwidth,
    height=0.01\textheight, 
    scale only axis,
    axis lines=none,
    ticks=none,
    xmin=0, xmax=1,
    ymin=0, ymax=1,
    legend columns=8,
    legend style={
        at={(0.5,0.5)},
        anchor=center,
        draw=black, 
        font=\tiny,
        fill=white, 
        /tikz/every even column/.append style={column sep=0.5cm}
    }
]
        \addlegendimage{mark=square*, color=blue}
        \addlegendentry{FedMPR}
        \addlegendimage{mark=o, color=red}
        \addlegendentry{FedSelect}
        \addlegendimage{mark=triangle*, color=green}
        \addlegendentry{FedDC}
        \addlegendimage{mark=star, color=orange}
        \addlegendentry{FedDyn}
        \addlegendimage{mark=x, color=purple}
        \addlegendentry{SCAFFOLD}
        \addlegendimage{mark=diamond*, color=cyan}
        \addlegendentry{FedAvg}
        \addlegendimage{mark=triangle, color=brown}
        \addlegendentry{FedProx}
    \end{axis}
    \end{tikzpicture}
\label{fig:plot1}
\caption{Top-1 accuracy on CIFAR-10 under covariate shifts with two clients and varying local sample sizes.}
\end{figure*}
Figures \ref{fig:fedmpr_tsne} illustrate the impact of the number of samples per class on representation learning with two clients under high and low covariate shift settings. As the number of samples per class increases, performance consistently improves in both scenarios, regardless of the degree of distributional shift.
\begin{table}[H]
\centering
\tiny
\setlength{\tabcolsep}{2pt}
\begin{tabular}{c|ccccccccc}
\toprule
\textbf{Dataset} & \textbf{$\alpha$} & \textbf{Clients} & \textbf{FedAvg} & \textbf{FedProx} & \textbf{FedDC} & \textbf{SCAFFOLD} & \textbf{FedDyn} & \textbf{FedSelect} & \textbf{FedMPR}  \\
\midrule \hline \hline
\multirow{4}{*}{CIFAR10} & \multirow{2}{*}{0.1} & 10  & 65.86 & 65.55 & 67.12 & 74.35 & \textbf{74.39} & 54.88 & 72.82  \\
                    &                      & 100 & 27.95 & 27.74 & 30.65 & 30.1 & 30.98 & 18.45 & \textbf{32.19}  \\
                    & \multirow{2}{*}{0.5} & 10  & 81.07 & 81.15 & 78.04 & 81.18 & 81.77 & 67.78 & \textbf{88.55}  \\
                    &                      & 100 & 39.62 & 40.01 & 40.91 & 39.08 & 41.58 & 19.16 & \textbf{49.91}  \\
\midrule \hline
\multirow{4}{*}{RAF-DB} & \multirow{2}{*}{0.1} & 10  & 41.82 & 41.36 & 39.77 & 40.35 & 43.68  & 64.34 & \textbf{67.73} \\
                    &                      & 100 & 17.63 & 18.29 & 18.38 & 18.58 & 19.04  & 39.86 & \textbf{44.85} \\
                    & \multirow{2}{*}{0.5} & 10  & 43.87 & 44.23 & 49.19 & 41.53 & 45.83  & 73.08 & \textbf{74.32} \\
                    &                      & 100 & 41.66 & 42.29 & 34.88 & 41.29 & 39.18  & 42.73 & \textbf{44.07}  \\
\midrule \hline
\multirow{4}{*}{CelebA} & \multirow{2}{*}{0.1} & 10  & 56.5 & 56.46 & 50.0 & 57.06 & 50.7  & 50.12 &\textbf{ 68.85 }  \\
                    &                      & 100 & 47.32 & 48.06 & 49.9 & 49.96 & 48.1 & \textbf{50.1} & 48.88  \\
                    & \multirow{2}{*}{0.5} & 10  & 83.76 & 80.8 & 55.13 & 80.62 & 50.62 & 85.32 & \textbf{90.05}  \\
                    &                      & 100 & 48.68 & 49.1 & 53.56 & 48.42 & 50.0 & \textbf{65.8} & 49.58  \\
\midrule \hline
\multirow{4}{*}{CelebA-Gender} & \multirow{2}{*}{0.1} & 10  & 70.2 & 71.0 & 52.2 & 66.19 & 65.4 & \textbf{72.9} & 65.11  \\
                    &                      & 100 & 50.8 & 54.88 & 49.0 & 42.6 & 49.2 & 49.04 & \textbf{58.36}  \\
                    & \multirow{2}{*}{0.5} & 10  & 79.2 & 79.8 & 57.8 & 78.8 & 71.4 & \textbf{77.58} & 76.5  \\
                    &                      & 100 & 53.6 & \textbf{62.8} & 53.4 & 59.2 & 58.2 & 51.0 & 52.76  \\
\bottomrule
\end{tabular}
\caption{Top-1 accuracy (\%) across datasets for varying $\alpha \in \{0.1, 0.5\}$ and client counts $\in \{10, 100\}$.}
\label{tab:results}
\end{table}
Pruning-based methods, such as \textsc{FedMPR}, demonstrate strong scalability under extreme data imbalance, outperforming baseline approaches in settings with 100 clients and a Dirichlet concentration parameter of $\alpha=0.1$ (Table~\ref{tab:results}). By promoting sparsity and applying regularisation during local training, these methods not only improve generalisation but also mitigate the adverse effects of model aggregation in non-IID environments. 
t-SNE plots in Figures \ref{fig:fedmpr_tsne} demonstrate that magnitude pruning effectively captures robust features across multiple datasets.
RAF-DB is inherently imbalanced without requiring synthetic non-IID partitioning, unlike other datasets that originally contain a uniform number of samples per class. Applying a Dirichlet distribution to such balanced datasets introduces severe non-IID conditions and significant class imbalance, making representation learning substantially more challenging. Nevertheless, our proposed method demonstrates strong robustness to this type of data heterogeneity.
\begin{figure*}[htpb]
    \centering
    \begin{minipage}[t]{0.24\textwidth}
        \centering
        \includegraphics[width=\linewidth]{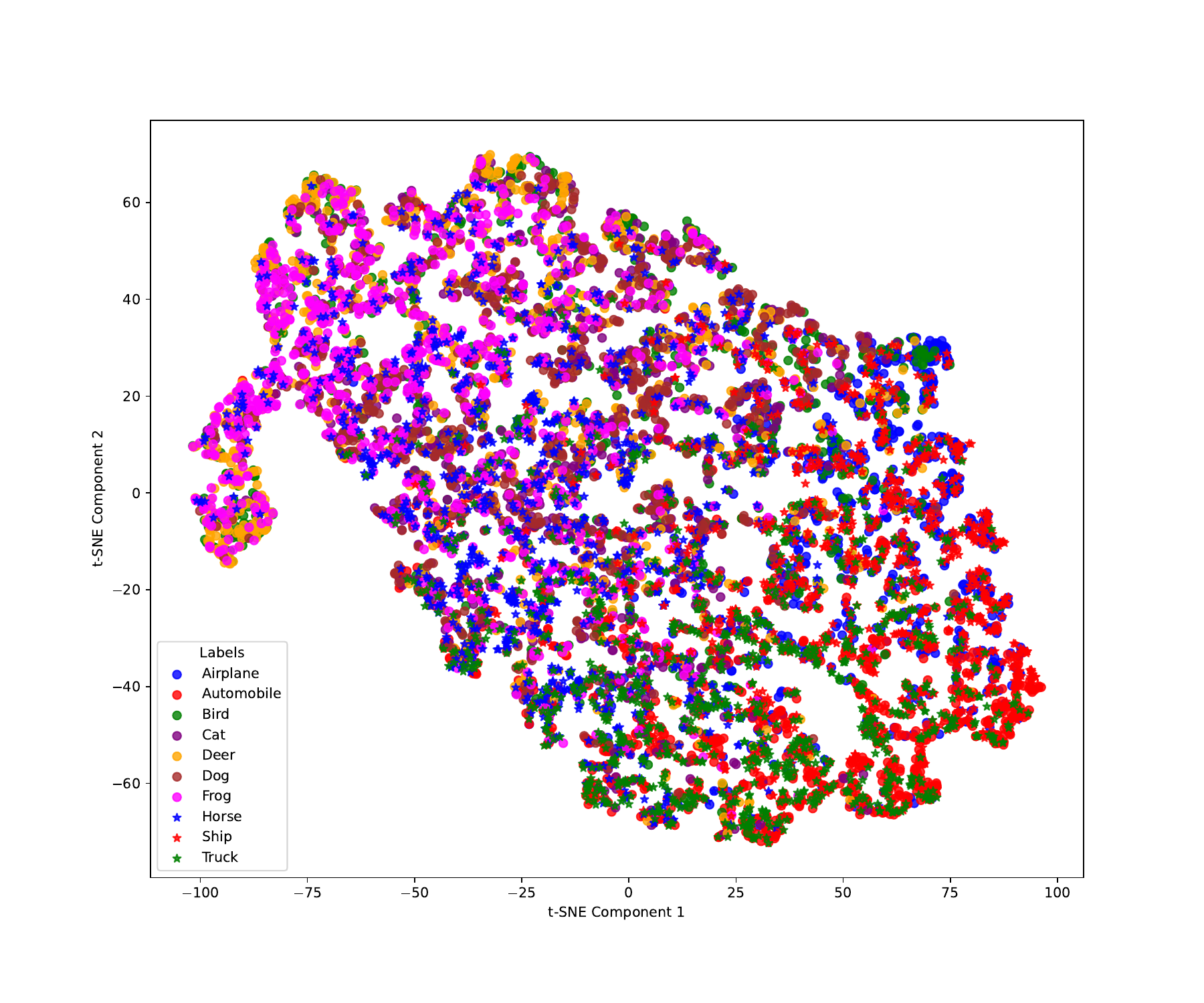}
        {\small CIFAR10}
    \end{minipage}
    \hfill
    \begin{minipage}[t]{0.24\textwidth}
        \centering
        \includegraphics[width=\linewidth]{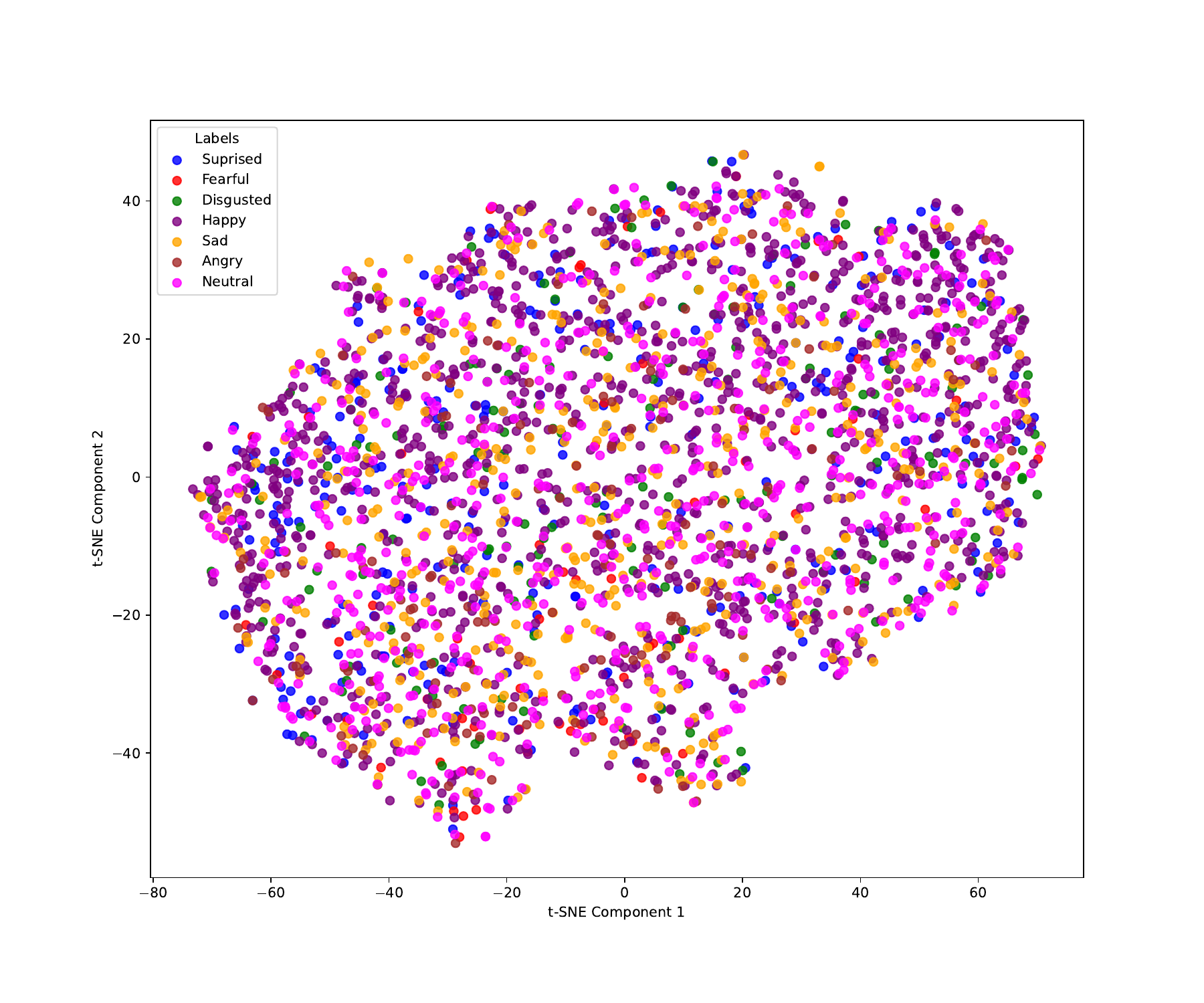}
        {\small RAF-DB}
    \end{minipage}
    \hfill
    \begin{minipage}[t]{0.24\textwidth}
        \centering
        \includegraphics[width=\linewidth]{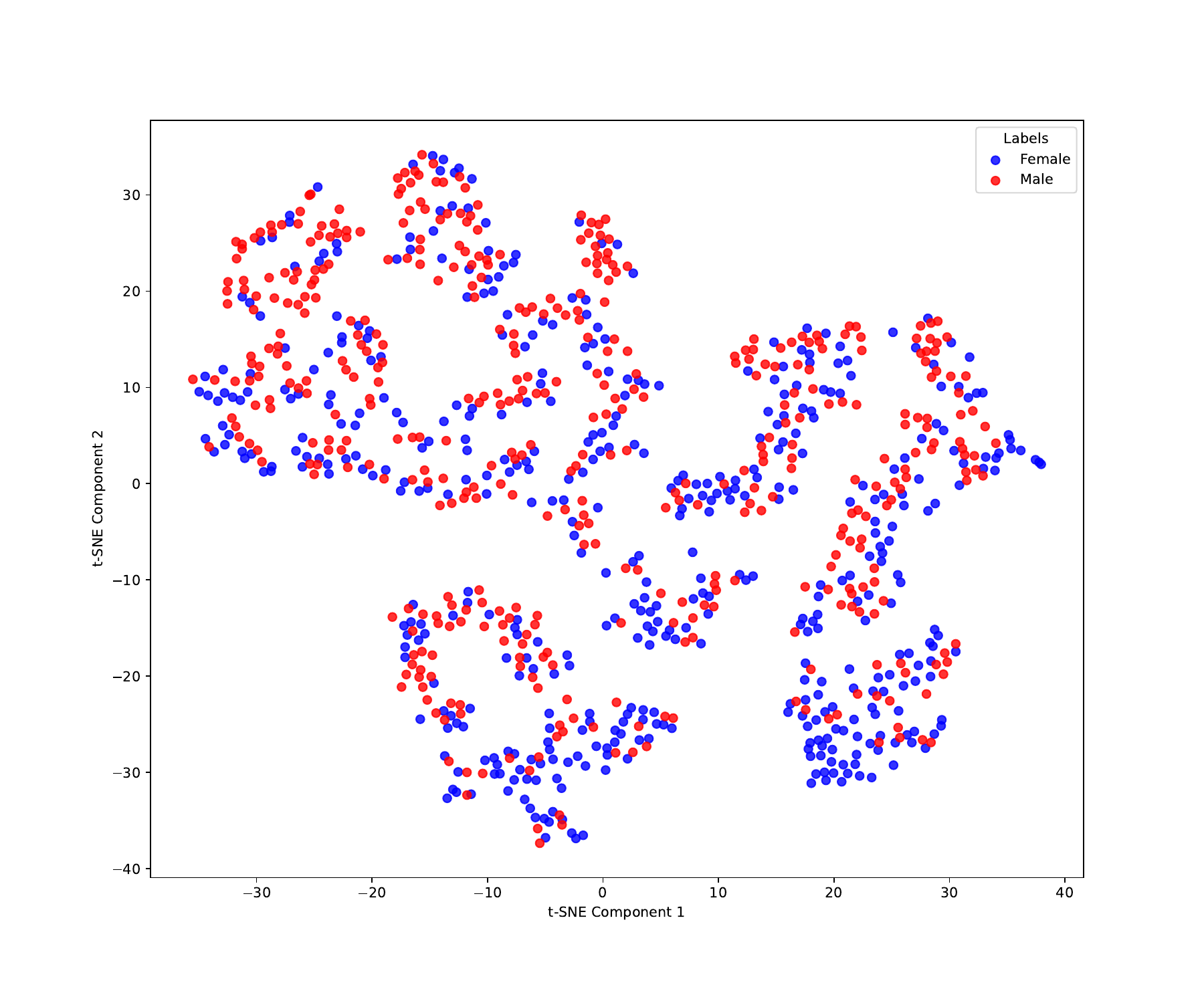}
        {\small CelebA-Gender}
    \end{minipage}
    \hfill
    \begin{minipage}[t]{0.24\textwidth}
        \centering
        \includegraphics[width=\linewidth]{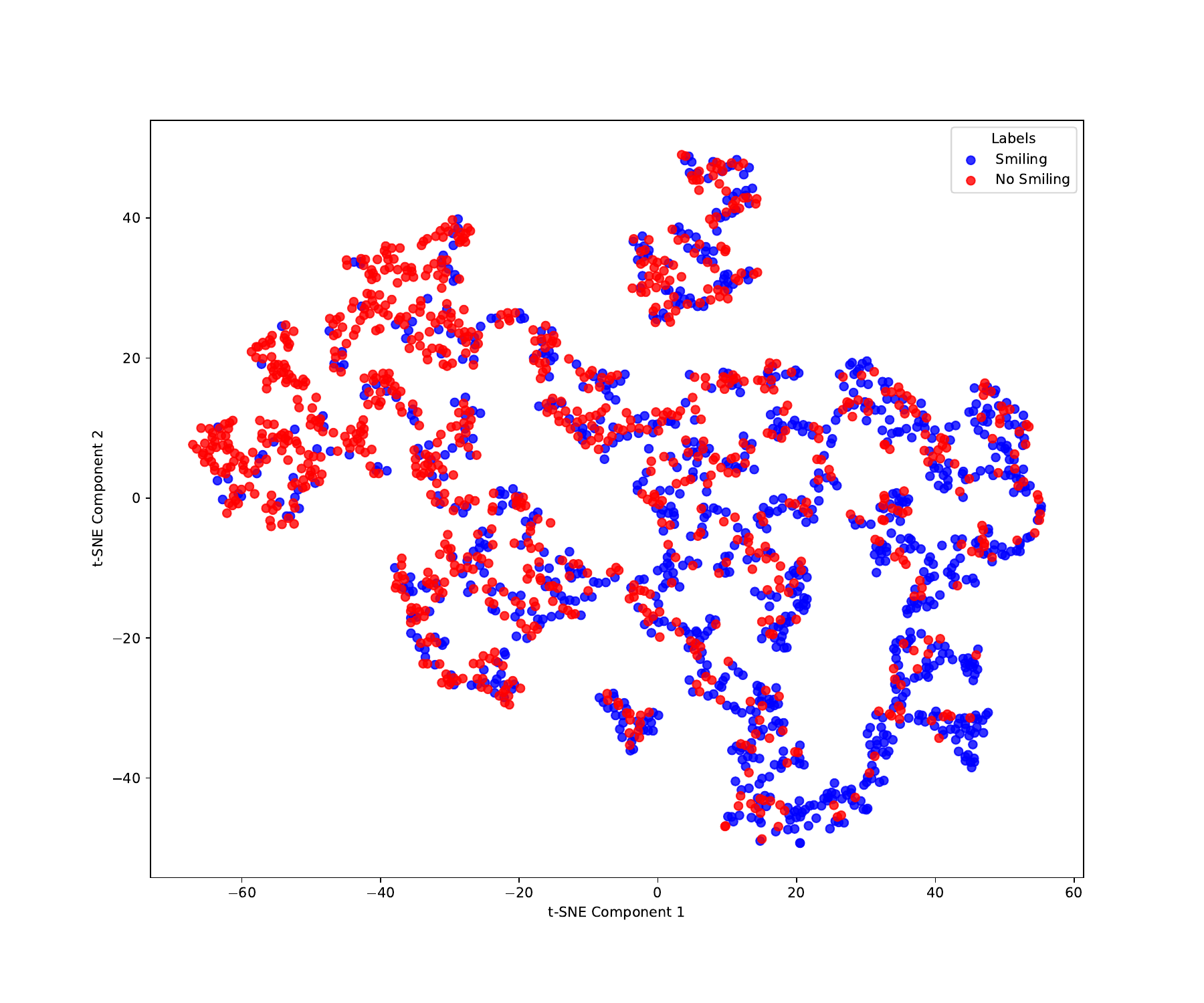}
        {\small CelebA}
    \end{minipage}
\vspace{0.5em} 
    
    \begin{minipage}[t]{0.24\textwidth}
        \centering
        \includegraphics[width=\linewidth]{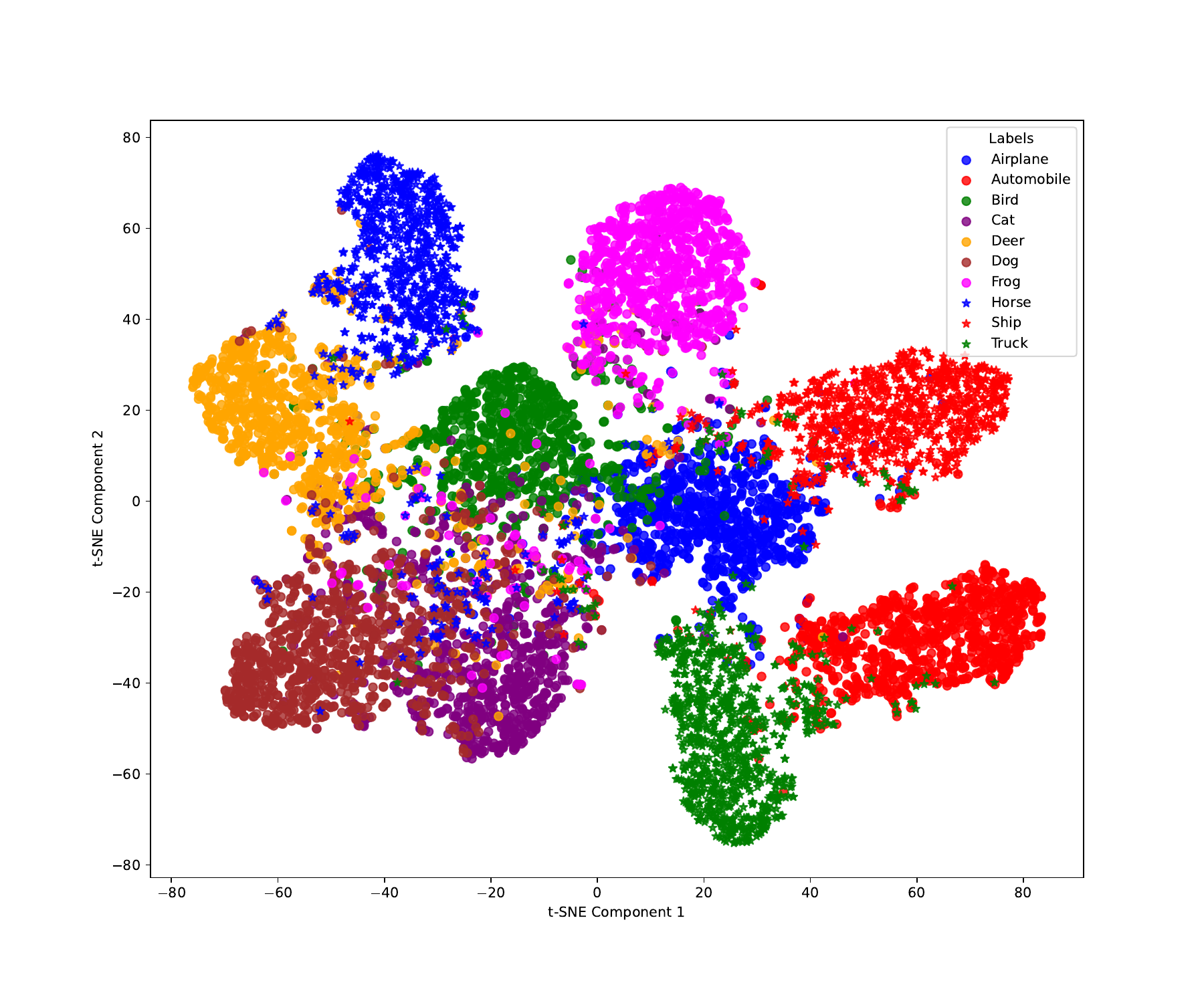}
        {\small CIFAR10}
    \end{minipage}
    \hfill
    \begin{minipage}[t]{0.24\textwidth}
        \centering
        \includegraphics[width=\linewidth]{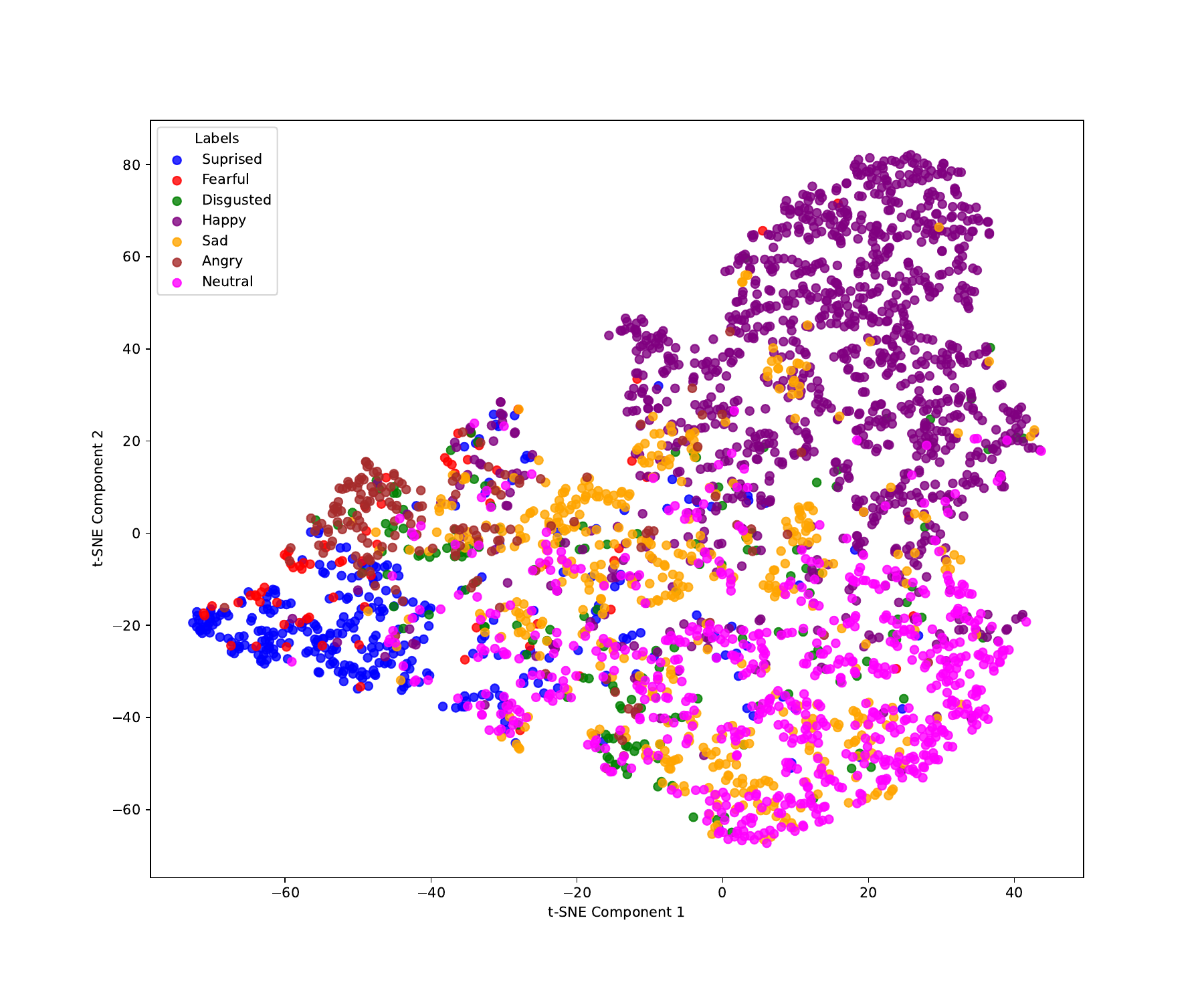}
        {\small RAF-DB}
    \end{minipage}
    \hfill
    \begin{minipage}[t]{0.24\textwidth}
        \centering
        \includegraphics[width=\linewidth]{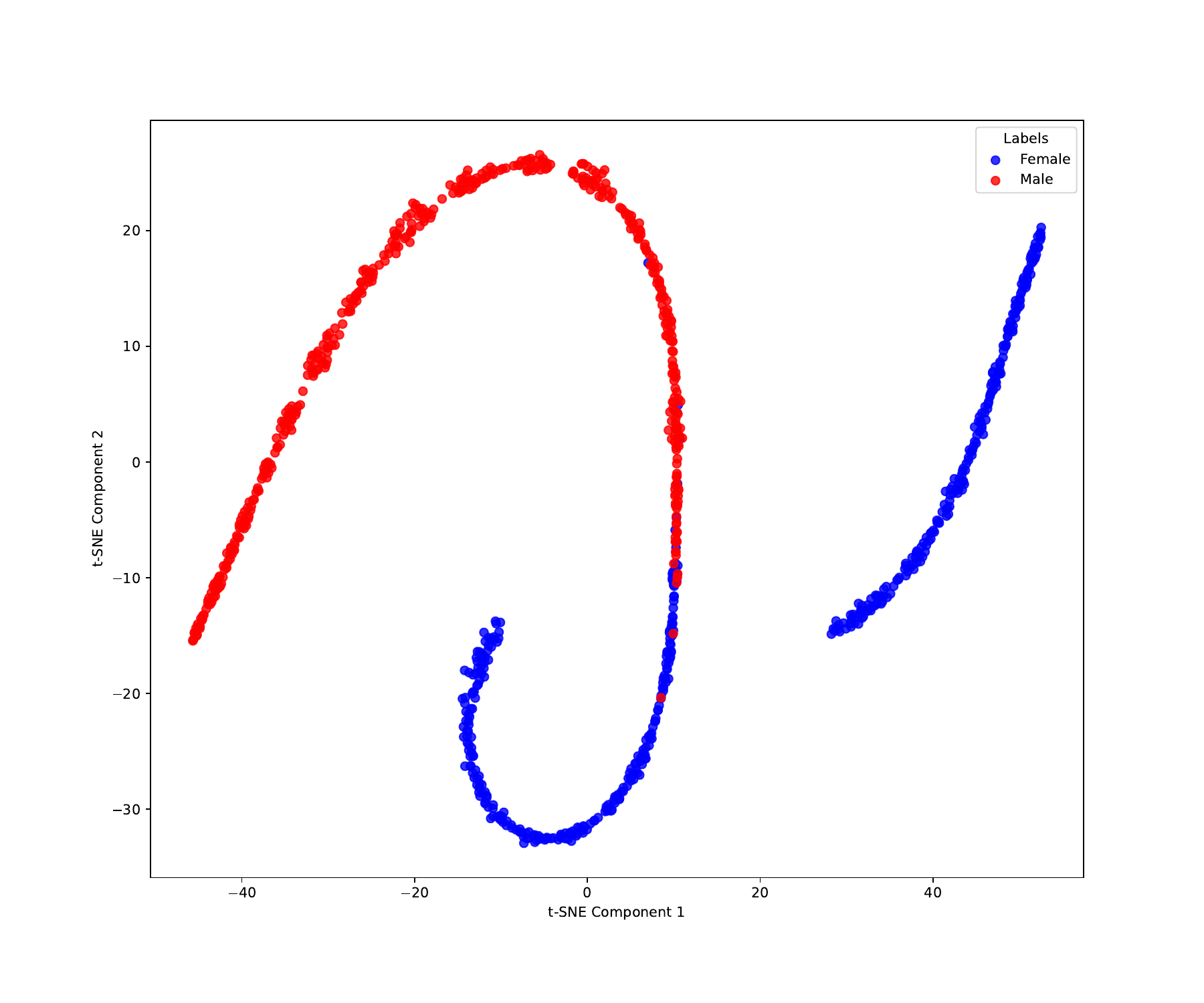}
        {\small CelebA-Gender}
    \end{minipage}
    \hfill
    \begin{minipage}[t]{0.24\textwidth}
        \centering
        \includegraphics[width=\linewidth]{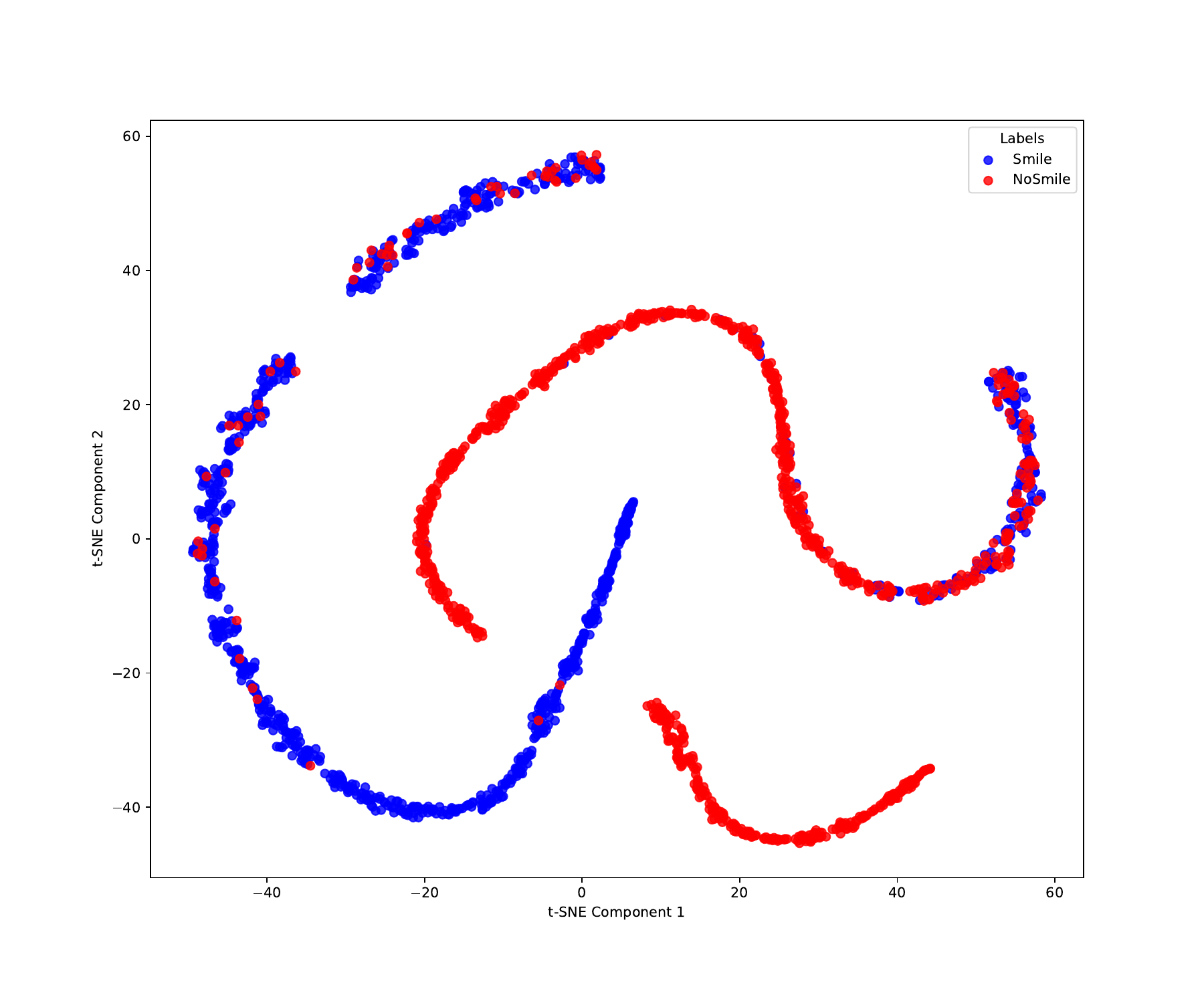}
        {\small CelebA}
    \end{minipage}
    \caption{t-SNE plots at the early stage of training (5th round embeddings) with $\alpha = 0.1$. The first row shows FedAvg results, while the second row depicts embeddings from \textsc{FedMPR}. }
    \label{fig:fedmpr_tsne}
\end{figure*}
\section{Ablation Study}
In order to deeply understand the respective importance of each components of the approach, we perform ablation experiments. In our proposed method, there are three main parts: pruning percentage (p), dropout ratio (d) and noise injection percentage (n).
Pruning percentages, dropout ratios and noise standard deviations were each examined within a range of 0.2 to 1. Table~\ref{tbl:as} shows the accuracy of the CIFAR10. 
The pruning has little impact on the low-CS however, provides a notable improvement in the high-CS. 
A small dropout ratio (d=0.2) provides a notable improvement in accuracy in both conditions, but larger values (increasing from 0.2 to 0.8) lead to a decrease in performance, especially in the high-CS (accuracy decreases from 78.2\% to 52.03\%). 
\textsc{FedMPR} denotes the optimal combination of parameters: a noise injection ratio of 0.4, a dropout ratio of 0.2 and a pruning ratio of 0.4. 
\begin{table}[htpb]
\footnotesize
    \begin{center}
        \begin{tabular}{c|c|c|c|c|c} 
            \textbf{Method} & \textbf{Prune} &  \textbf{Dropout} & \textbf{Noise}  & \textbf{low-CS} & \textbf{high-CS}  \\ [0.5ex] 
            \hline  FL (Baseline) & \xmark  & \xmark & \xmark & 82.64 & 72.67
            \\
             \hline  FL (p=0.2) & \cmark  & \xmark & \xmark & 81.84 & 71.25
            \\
             \hline  FL (p=0.4) & \cmark  & \xmark & \xmark & 81.72 & 75.63
            \\
              \hline  FL (d=0.2) & \xmark  & \cmark & \xmark & 86.18 & 78.20
            \\
             \hline  FL (d=0.8) & \xmark  & \cmark & \xmark & 80.36 & 52.03
            \\
             \hline  FL (n=0.2) & \xmark  & \xmark & \cmark & 83.10 & 74.83
            \\
            \hline  FL (n=0.4) & \xmark  & \xmark & \cmark & 84.26 & 75.76
            \\
              \hline  \thead{FL (p=0.4),\\ (d=0.2)} & \cmark  & \cmark & \xmark & 79.82 &  71.43
              \\
              \hline  \thead{FL (d=0.2),\\ (n=0.4)} & \xmark  & \cmark & \cmark & 77.22  & 70.09
            \\
              \hline  \thead{FL (p=0.4), \\ (n=0.4)} & \cmark  & \xmark & \cmark & 77.35 & 68.92
            \\ 
             \hline \thead{ \textsc{FedMPR} \\ (p=0.4), \\ (d=0.2), (n=0.4) }  & \cmark  & \cmark & \cmark & \textbf{87.89} & \textbf{80.57}            
        \end{tabular}
    \caption{Impact of each component on FL for low-CS and high-CS on top-1 accuracy on CIFAR10 with best accuracy under many runs.}
    \label{tbl:as}
    \end{center}
\end{table}
\section{Conclusion}
Federated Learning (FL) enables decentralised model training without sharing raw data, but suffers from severe covariate shifts induced by inconsistent local updates across heterogeneous clients. We present \textsc{FedMPR}, a pruning-based FL framework that integrates iterative unstructured pruning with dropout and noise injection to enhance robustness against data heterogeneity, even when only a small subset of clients participates. Extensive experiments on standard benchmarks show that \textsc{FedMPR} consistently outperforms state-of-the-art baselines. To enable controlled evaluation, we introduce CelebA-Gender, a benchmark dataset specifically designed to isolate covariate shift by altering within-class distributions while maintaining class balance. Results demonstrate that pruning-based regularisation effectively mitigates distributional shifts without requiring explicit personalisation strategies. Moreover, in both small and large client regimes with high data heterogeneity or when accessible clients exhibit higher data diversity, our method further improves FL performance. Future work will extend \textsc{FedMPR} to dynamic, large-scale, and personalised FL methods comparison with full client participation.

\begin{thebibliography}{00}
\bibitem{fedavg} 
B. McMahan, E. Moore, D. Ramage, S. Hampson, and B. A. y Arcas, "Communication-efficient learning of deep networks from decentralized data," in \textit{Proc. Artificial Intelligence and Statistics (AISTATS)}, 2017, pp. 1273--1282.

\bibitem{fedprox}
P. Sharma, R. Panda, G. Joshi, and P. Varshney, "Federated minimax optimization: Improved convergence analyses and algorithms," in \textit{Proc. International Conference on Machine Learning (ICML)}, 2022, pp. 19683--19730.

\bibitem{gao2022feddc} 
L. Gao, H. Fu, L. Li, Y. Chen, M. Xu, and C.-Z. Xu, "FedDC: Federated learning with non-IID data via local drift decoupling and correction," in \textit{Proc. IEEE/CVF Conf. Computer Vision and Pattern Recognition (CVPR)}, 2022, pp. 10112--10121.

\bibitem{liu2015deep} 
Z. Liu, P. Luo, X. Wang, and X. Tang, "Deep learning face attributes in the wild," in \textit{Proc. IEEE Int. Conf. Computer Vision (ICCV)}, 2015, pp. 3730--3738.

\bibitem{caldas2018leaf}
S. Caldas, S. M. K. Duddu, P. Wu, T. Li, J. Konečný, H. B. McMahan, V. Smith, and A. Talwalkar, "Leaf: A benchmark for federated settings," \textit{arXiv preprint arXiv:1812.01097}, 2018.

\bibitem{zhu2021federated} 
H. Zhu, J. Xu, S. Liu, and Y. Jin, "Federated learning on non-IID data: A survey," \textit{Neurocomputing}, vol. 465, pp. 371--390, 2021.

\bibitem{oh2021fedbabu} 
J. Oh, S. Kim, and S. Yun, "FedBabu: Towards enhanced representation for federated image classification," \textit{International Conference on Learning Representations (ICLR)}, 2021.

\bibitem{tamirisa2024fedselect} 
R. Tamirisa, C. Xie, W. Bao, A. Zhou, R. Arel, and A. Shamsian, "FedSelect: Personalized federated learning with customized selection of parameters for fine-tuning," in \textit{Proc. IEEE/CVF Conf. Computer Vision and Pattern Recognition (CVPR)}, 2024, pp. 23985--23994.

\bibitem{li2021model}
Q. Li, B. He, and D. Song, "Model-contrastive federated learning," in \textit{Proc. IEEE/CVF Conf. Comput. Vis. Pattern Recognit.}, 2021, pp. 10713--10722.

\bibitem{tan2022towards}
A. Z. Tan, H. Yu, L. Cui, and Q. Yang, "Towards personalized federated learning," \textit{IEEE Transactions on Neural Networks and Learning Systems}, vol. 34, no. 12, pp. 9587--9603, 2022.

\bibitem{jia2024dapperfl}
Y. Jia, X. Zhang, H. Hu, K.-K. R. Choo, L. Qi, X. Xu, A. Beheshti, and W. Dou, "DapperFL: Domain adaptive federated learning with model fusion pruning for edge devices," in \textit{Advances in Neural Information Processing Systems}, vol. 37, pp. 13099--13123, 2024.

\bibitem{luo2021no}
M. Luo, F. Chen, D. Hu, Y. Zhang, J. Liang, and J. Feng, "No fear of heterogeneity: Classifier calibration for federated learning with non-iid data," \textit{Advances in Neural Information Processing Systems}, vol. 34, pp. 5972--5984, 2021.

\bibitem{dai2023tackling} 
Y. Dai, Z. Chen, J. Li, S. Heinecke, L. Sun, and R. Xu, "Tackling data heterogeneity in federated learning with class prototypes," in \textit{Proc. AAAI Conf. Artificial Intelligence}, vol. 37, 2023, pp. 7314--7322.

\bibitem{huang2023rethinking} 
W. Huang, M. Ye, Z. Shi, H. Li, and B. Du, "Rethinking federated learning with domain shift: A prototype view," in \textit{Proc. IEEE/CVF Conf. Computer Vision and Pattern Recognition (CVPR)}, 2023, pp. 16312--16322.

\bibitem{li2022federated}
Q. Li, Y. Diao, Q. Chen, and B. He, "Federated learning on non-iid data silos: An experimental study," in \textit{2022 IEEE 38th International Conference on Data Engineering (ICDE)}, pp. 965--978, 2022.

\bibitem{collins2021exploiting}
L. Collins, H. Hassani, A. Mokhtari, and S. Shakkottai, "Exploiting shared representations for personalized federated learning," in \textit{Proc. Int. Conf. Mach. Learn.}, 2021, pp. 2089--2099.

\bibitem{karimireddy2019scaffold} 
S. P. Karimireddy, S. Kale, M. Mohri, S. J. Reddi, S. U. Stich, and A. T. Suresh, "SCAFFOLD: Stochastic controlled averaging for on-device federated learning," \textit{arXiv preprint arXiv:1910.06378}, vol. 2, no. 6, 2019.

\bibitem{lee2024regularizing}
G. Lee and D. Choi, "Regularizing and aggregating clients with class distribution for personalized federated learning," \textit{arXiv preprint arXiv:2406.07800}, 2024.

\bibitem{wen2022federated}
D. Wen, K.-J. Jeon, and K. Huang, "Federated dropout A simple approach for enabling federated learning on resource constrained devices," \textit{IEEE Wireless Commun. Lett.}, vol. 11, no. 5, pp. 923--927, 2022.

\bibitem{frankle2018lottery}
J. Frankle and M. Carbin, "The lottery ticket hypothesis: Finding sparse, trainable neural networks," in \textit{International Conference on Learning Representations}, 2019.

\bibitem{han2015learning}
S. Han, J. Pool, J. Tran, and W. Dally, "Learning both weights and connections for efficient neural network," in \textit{Advances in Neural Information Processing Systems}, vol. 28, 2015.

\bibitem{feddyn} 
D. Acar, Y. Zhao, R. M. Navarro, M. Mattina, P. N. Whatmough, and V. Saligrama, "Federated learning based on dynamic regularization," \textit{arXiv preprint arXiv:2111.04263}, 2021.

\bibitem{peng2019federated} 
X. Peng, Z. Huang, Y. Zhu, and K. Saenko, "Federated adversarial domain adaptation," \textit{arXiv preprint arXiv:1911.02054}, 2019.

\bibitem{lu2023feddad} 
P. J. Lu, C.-Y. Jui, and J.-H. Chuang, "FedDAD: Federated domain adaptation for object detection," \textit{IEEE Access}, vol. 11, pp. 51320--51330, 2023.

\bibitem{yao2022federated} 
C.-H. Yao, B. Gong, H. Qi, Y. Cui, Y. Zhu, and M.-H. Yang, "Federated multi-target domain adaptation," in \textit{Proc. IEEE/CVF Winter Conf. Applications of Computer Vision (WACV)}, 2022, pp. 1424--1433.

\bibitem{jayasumana2024rethinking} 
S. Jayasumana, S. Ramalingam, A. Veit, D. Glasner, A. Chakrabarti, and S. Kumar, "Rethinking FID: Towards a better evaluation metric for image generation," in \textit{Proc. IEEE/CVF Conf. Computer Vision and Pattern Recognition (CVPR)}, 2024, pp. 9307--9315.

\bibitem{mendieta2022local} 
M. Mendieta, T. Yang, P. Wang, M. Lee, Z. Ding, and C. Chen, "Local learning matters: Rethinking data heterogeneity in federated learning," in \textit{Proc. IEEE/CVF Conf. Computer Vision and Pattern Recognition (CVPR)}, 2022, pp. 8397--8406.

\bibitem{mora2024enhancing} 
A. Mora, A. Bujari, and P. Bellavista, "Enhancing generalization in federated learning with heterogeneous data: A comparative literature review," \textit{Future Generation Computer Systems}, 2024.

\bibitem{liang2020think} 
P. P. Liang, T. Liu, Z. Liu, N. B. Allen, R. P. Auerbach, D. Brent, R. Salakhutdinov, and L.-P. Morency, "Think locally, act globally: Federated learning with local and global representations," \textit{arXiv preprint arXiv:2001.01523}, 2020.

\bibitem{fednas} 
C. He, M. Annavaram, and S. Avestimehr, "Towards non-IID and invisible data with FedNAS: Federated deep learning via neural architecture search," \textit{arXiv preprint arXiv:2004.08546}, 2020.

\bibitem{krizhevsky2009learning} 
A. Krizhevsky, "Learning multiple layers of features from tiny images," Master's thesis, University of Toronto, 2009.

\bibitem{lecun1998gradient} 
Y. LeCun, L. Bottou, Y. Bengio, and P. Haffner, "Gradient-based learning applied to document recognition," \textit{Proc. IEEE}, vol. 86, no. 11, pp. 2278--2324, 1998.


\bibitem{howard2020fastai}
J. Howard and S. Gugger, "Fastai: a layered API for deep learning," \textit{Information}, vol. 11, no. 2, p. 108, 2020.

\bibitem{netzer2011reading}
Y. Netzer, T. Wang, A. Coates, A. Bissacco, B. Wu, A. Y. Ng, et al., "Reading digits in natural images with unsupervised feature learning," in \textit{NIPS workshop on deep learning and unsupervised feature learning}, Granada, 2011, p. 4.

\bibitem{xiao2017fashion}
H. Xiao, K. Rasul, and R. Vollgraf, "Fashion-MNIST: a novel image dataset for benchmarking machine learning algorithms," \textit{arXiv preprint arXiv:1708.07747}, 2017.

\bibitem{srivastava2014dropout}
N. Srivastava, G. Hinton, A. Krizhevsky, I. Sutskever, and R. Salakhutdinov, "Dropout: a simple way to prevent neural networks from overfitting," \textit{The Journal of Machine Learning Research}, vol. 15, no. 1, pp. 1929--1958, 2014.

\bibitem{fallah2020personalized}
A. Fallah, A. Mokhtari, and A. Ozdaglar, "Personalized federated learning with theoretical guarantees: A model-agnostic meta-learning approach," \textit{Advances in Neural Information Processing Systems}, vol. 33, pp. 3557--3568, 2020.

\bibitem{chen2018federated}
F. Chen, M. Luo, Z. Dong, Z. Li, and X. He, "Federated meta-learning with fast convergence and efficient communication," \textit{arXiv preprint arXiv:1802.07876}, 2018.

\bibitem{jiang2019fantastic}
Y. Jiang, B. Neyshabur, H. Mobahi, D. Krishnan, and S. Bengio, "Fantastic generalization measures and where to find them," \textit{arXiv preprint arXiv:1912.02178}, 2019.

\bibitem{kundu2022robustness}
A. Kundu, P. Yu, L. Wynter, and S. H. Lim, "Robustness and personalization in federated learning: A unified approach via regularization," in \textit{2022 IEEE International Conference on Edge Computing and Communications (EDGE)}, pp. 1--11, 2022.

\bibitem{lai2024personalized}
Y.-H. Lai, S.-Y. Chen, W.-C. Chou, H.-Y. Hsu, and H.-C. Chao, "Personalized federated learning with adaptive feature extraction and category prediction in non-IID datasets," \textit{Future Internet}, vol. 16, no. 3, p. 95, 2024.

\bibitem{li2017reliable}
S. Li, W. Deng, and J. Du, "Reliable crowdsourcing and deep locality-preserving learning for expression recognition in the wild," in \textit{Proc. IEEE Conf. Comput. Vis. Pattern Recognit.}, 2017, pp. 2852--2861.


\bibitem{pillutla2022federated}
K. Pillutla, K. Malik, A.-R. Mohamed, M. Rabbat, M. Sanjabi, and L. Xiao, "Federated learning with partial model personalization," in \textit{Proc. Int. Conf. Mach. Learn.}, 2022, pp. 17716--17758.

\bibitem{jiang2022model}
Y. Jiang, S. Wang, V. Valls, B. J. Ko, W.-H. Lee, K. K. Leung, and L. Tassiulas, "Model pruning enables efficient federated learning on edge devices," \textit{IEEE Trans. Neural Netw. Learn. Syst.}, vol. 34, no. 12, pp. 10374--10386, 2022.

\bibitem{li2020lotteryfl}
A. Li, J. Sun, B. Wang, L. Duan, S. Li, Y. Chen, and H. Li, "Lotteryfl: Personalized and communication-efficient federated learning with lottery ticket hypothesis on non-iid datasets," \textit{arXiv preprint arXiv:2008.03371}, 2020.

\bibitem{heusel2017gans}
M. Heusel, H. Ramsauer, T. Unterthiner, B. Nessler, and S. Hochreiter, "Gans trained by a two time-scale update rule converge to a local nash equilibrium," in \textit{Advances in Neural Information Processing Systems}, vol. 30, 2017.

\end{thebibliography}

\end{document}